\documentclass[10pt,twocolumn,letterpaper]{article}

\usepackage{iccv}              % To produce the CAMERA-READY version
\definecolor{iccvblue}{rgb}{0.21,0.49,0.74}
\usepackage[pagebackref,breaklinks,colorlinks,allcolors=iccvblue]{hyperref}
\usepackage{amsmath}

\usepackage{amsmath}
\usepackage[normalem]{ulem}
\def\paperID{11460} % *** Enter the Paper ID here
\def\confName{ICCV}
\def\confYear{2025}

\title{\textcolor{blue}{DEL}: \textcolor{blue}{D}ense \textcolor{blue}{E}vent \textcolor{blue}{L}ocalization for Multi-modal Audio-Visual Understanding}

\author{Mona Ahmadian\\
University of Surrey\\
{\tt\small m.ahmadian@surrey.ac.uk}
\and
Amir Shirian\\
JPMorgan Chase\\
{\tt\small amirdonte15@gmail.com}
\and
Frank Guerin\\
University of Surrey\\
{\tt\small f.guerin@surrey.ac.uk}
\and
Andrew Gilbert\\
University of Surrey\\
{\tt\small a.gilbert@surrey.ac.uk}
}
\begin{document}
\maketitle
\begin{abstract}
Real-world videos often contain overlapping events and complex temporal dependencies, making multimodal interaction modeling particularly challenging. We introduce \textbf{DEL}, a framework for dense semantic action localization, aiming to accurately detect and classify multiple actions at fine-grained temporal resolutions in long untrimmed videos. DEL consists of two key modules: the alignment of audio and visual features that leverage masked self-attention to enhance intra-mode consistency and a multimodal interaction refinement module that models cross-modal dependencies across multiple scales, enabling high-level semantics and fine-grained details. Our method achieves state-of-the-art performance on multiple real-world Temporal Action Localization (TAL) datasets, UnAV-100, THUMOS14, ActivityNet 1.3, and EPIC-Kitchens-100, surpassing previous approaches with notable average mAP gains of +3.3\%, +2.6\%, +1.2\%, +1.7\% (verb), and +1.4\% (noun), respectively. 
%These results demonstrate the effectiveness of our approach and highlight the importance of multi-scale cross-modal perception and dependency modeling in advancing audio-visual scene understanding. 
The source code will be made publicly available.\end{abstract}    
\section{Introduction}
\label{sec:intro}

Temporal action localization (TAL) is a crucial and challenging task in video understanding, focusing on identifying the temporal boundaries of action instances and classifying them within untrimmed videos~\cite{vahdani2022deep}. Understanding real-world scenes and events is inherently a multimodal perception process, integrating visual and auditory cues to achieve comprehensive video understanding~\cite{jiao2024causal, nagrani2021attention,owens2018audio}. For instance, distinguishing between a person speaking and merely mouthing words is difficult with visual information alone, as both actions involve similar lip movements. Incorporating audio cues, however, helps resolve this ambiguity by detecting the presence or absence of vocal sounds.
% However, accurately modeling these multimodal interactions remains a significant challenge, especially in real-world scenarios where videos often contain overlapping, concurrent events and complex temporal dependencies.
Despite the complementary nature of visual and audio information, modeling multimodal interactions remains challenging due to modality misalignment, the length of the actions, and the complex interplay of cross-modal dependencies—particularly in videos with overlapping, concurrent events~\cite{tian2018audio}.

\begin{figure}[tb]
  \centering
\includegraphics[trim={1.8cm 3cm 0cm 0cm},width=1.109\linewidth]{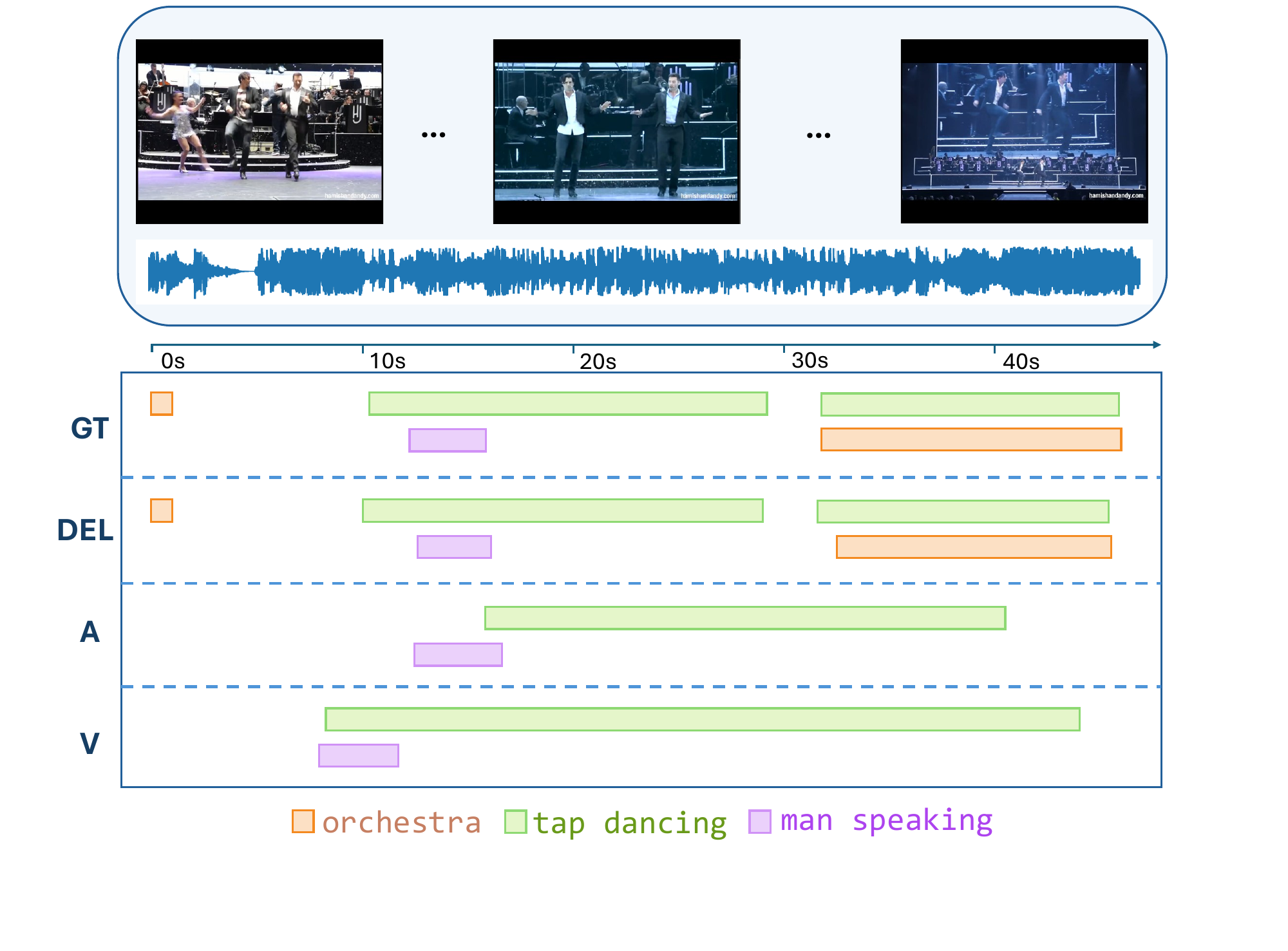}
  \vspace{-10pt}
  
  \caption{Real-world videos contain overlapping events of varying durations, making precise localization challenging. The image presents ground-truth (GT) annotations alongside the predictions of DEL, an audio-only model (A), and a visual-only model (V). The unimodal models struggle with a certain category. In contrast, \textbf{DEL} effectively detects and classifies both short and long-duration events, even when they co-occur, demonstrating its superior multimodal fusion capability.}
  \label{fig:teaser}
\end{figure}

Prior efforts have made progress in Audio-Visual Event Localization (AVE) for trimmed videos, where each video typically contains a single, isolated audio-visual event~\cite{gemmeke2017audio, venkatesh2022you, prashanth2024review}. 
%—these settings do not reflect the complexity of untrimmed videos lasting typically up to \AG{??} minutes in length, where multiple events with varying durations frequently co-occur~\cite{geng2023dense, idrees2017thumos, caba2015activitynet, damen2022rescaling}.
%%FG cut above because it says the same information as the next sentence
Unlike AVE, dense localization of audio-visual events requires identifying and recognizing all events within an untrimmed video, capturing short and long-duration interactions that may overlap~\cite{geng2023dense, idrees2017thumos, caba2015activitynet, damen2022rescaling}. Recent advances have demonstrated the effectiveness of transformer networks and Feature Pyramid Networks (FPN) for TAL~\cite{yang2020temporal, zhang2022actionformer, weng2022efficient, shi2023temporal}, significantly improving performance by leveraging multi-resolution visual features. However, those approaches overlook the role of audio.
%Multimodal audio-visual approaches face challenges due to modality misalignment, and the need to fuse audio features across different temporal scales. %This gap raises the key research question:

%\textbf{How can audio-visual feature representations effectively enhance the interpretation of long, untrimmed videos?}

A persistent challenge in audio-visual event localization is the effective extraction and fusion of information across modalities, especially when multiple events co-occur. Traditional approaches either process audio and visual streams independently or rely on late fusion, limiting their ability to model fine-grained temporal dependencies. Moreover, most multi-modal methods employ pretrained models to extract features, which, while beneficial for representation learning, often lead to misaligned features due to differences in pretraining objectives and domain shifts between audio and visual modalities. Additionally, contrastive learning techniques often focus on instance-level alignment while neglecting intra-video relationships such as temporal coherence and cross-event correlations, crucial for distinguishing similar events occurring at different times.
 % While audio-visual data provides rich contextual cues, effectively leveraging these multimodal representations for improved video interpretation remains an open problem. 
 % Existing approaches often treat audio and visual modalities separately or rely on late fusion strategies, limiting their ability to capture fine-grained temporal dependencies.

%In this work, we address the task of dense semantic action localization in untrimmed videos, which requires precise temporal alignment and fine-grained spatial understanding across multiple audio-visual events. Existing approaches often model audio and visual modalities independently, limiting their ability to capture the intricate dependencies and correlations between them. Understanding events in untrimmed videos requires effectively integrating visual and auditory cues, ensuring that the temporal structure and cross-modal dependencies are captured accurately. Traditional methods often struggle with misaligning audio and visual features, leading to poor localization and classification performance. Moreover, existing contrastive learning approaches focus on instance-level alignment but fail to capture the fine-grained relationships within each video, which is essential for distinguishing similar events occurring at different times. To overcome these limitations, we propose a novel framework that effectively integrates temporal coherence and spatial detail through multimodal fusion, leveraging the complementary strengths of audio and visual features.

To address this challenge, we propose \textbf{DEL}, a novel framework designed to model cross-modal dependencies while explicitly preserving fine-grained temporal structure. Our approach enables precise temporal alignment and fine-grained spatial understanding across multiple audio-visual events. DEL uses a multi-scale fusion strategy, ensuring robust event localization even in densely overlapping scenarios. We incorporate two key modules into our transformer-based model: 1) a \textbf{multimodal adaptive attention mechanism} using a masked self-attention mechanism that ensures temporal coherence and intra-modal consistency within the audio-visual information. 2) a \textbf{path aggregation network} that introduces multiple temporal resolutions to capture fine-grained and high-level semantics in the sequences. 

Additionally, we introduce an intra-sample contrastive loss to refine feature discrimination within a single data mode and an inter-sample contrastive loss for better alignment of audio-visual representations across different samples. %Our novel score-based contrastive pair selection helps the model dynamically select hard negatives based on event confidence, 
Instead of manually defining positive and negative samples, our method introduces a feature scoring mechanism that enables the model to automatically assign confidence scores and select contrastive pairs during training. This enhances feature discrimination and cross-modal coherence for precise event localization.
We demonstrate state-of-the-art performance on multiple benchmarks, including  UnAV-100~\cite{geng2023dense}, THUMOS14~\cite{idrees2017thumos}, ActivityNet 1.3~\cite{caba2015activitynet}, and EPIC-Kitchens-100~\cite{damen2022rescaling}, achieving significant improvements over prior methods.

\section{Related Works}
\label{sec:rworks}

\subsection{Single-Modality Temporal Localization Tasks}

Deep learning has significantly advanced temporal action localization (TAL), enabling the detection and classification of actions in untrimmed videos. TAL methods are generally categorized into two-stage and single-stage approaches. Two-stage methods generate action proposals with confidence scores before refining and classifying them~\cite{lin2019bmn,lin2018bsn}. In contrast, single-stage approaches directly localize actions in a single pass, eliminating the need for proposal generation. Our work follows a single-stage TAL. 

These methods can be further classified into anchor-based~\cite{chao2018rethinking,long2019gaussian,zhao2021video,wang2022rcl} and anchor-free approaches~\cite{lin2021learning, yang2020revisiting,nag2022proposal,xia2022dual,xia2022learning}. Anchor-based methods employ predefined multi-scale temporal anchors to detect potential action segments, refining their boundaries through regression techniques. Conversely, anchor-free methods predict action boundaries directly by modeling temporal dependencies or regressing distances to action start and end points.

Recent advances integrate graph neural networks (GNNs)~\cite{zeng2019graph,yang2020revisiting,xu2020g} and transformers~\cite{wang2021temporal,tan2021relaxed,chang2022augmented} to enhance temporal modeling. Transformers, in particular, have demonstrated superior performance by leveraging self-attention mechanisms to capture long-range dependencies. Inspired by progress in object detection~\cite{redmon2016you,cheng2024yolo} and saliency detection~\cite{lin2021learning}, recent state-of-the-art approaches incorporate transformer-based feature pyramid networks (FPNs)~\cite{cheng2022tallformer,weng2022efficient,zhang2022actionformer,shi2023temporal}. These networks effectively combine multi-scale feature representation with classification and regression heads, improving accuracy and efficiency over conventional methods. 

\begin{figure*}[t]
    \centering
    \includegraphics[trim={3.5cm 4.4cm 0cm 2cm}, height=6cm]{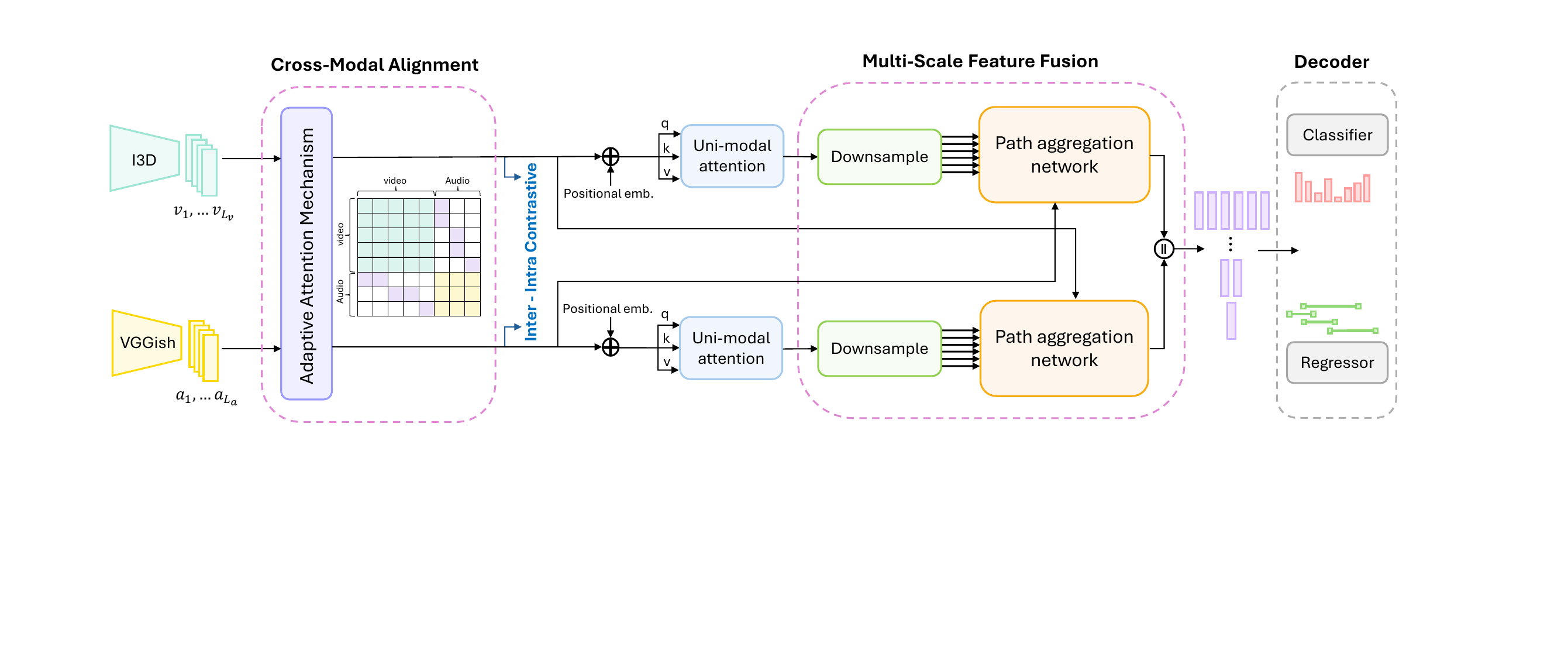}
      \vspace{-40pt}

    \caption{Overview of our proposed \textbf{DEL} framework. 
    Our model integrates \textbf{(1)} an \textit{adaptive attention mechanism} for aligning audio and visual features, \textbf{(2)} \textit{inter- and intra-sample contrastive learning} to enhance event discrimination, and \textbf{(3)} a \textit{multi-scale path aggregation network} for feature fusion. $\mathbin{\|}$ represents the concatenation operation. 
    %DEL efficiently localizes fine-grained and overlapping events in untrimmed videos, leveraging cross-modal dependencies for improved accuracy.
    }
    \label{fig:del_model}
\end{figure*}

%-------------------------------------------------------------------------
\subsection{Multimodal Audio-Visual Event Localization}

While joint audio-visual representation learning has been extensively studied in video retrieval and classification~\cite{kazakos2019epic,xiao2020audiovisual,alcazar2021maas}, its application in TAL is less explored, primarily due to the challenges of modality misalignment and the need for fine-grained temporal modeling.
Most existing methods assume that the audio and visual events are perfectly aligned~\cite{tian2018audio,bagchi2021hear,yu2021mpn}, limiting their effectiveness in real-world settings where events may be asynchronous or overlapping.  As highlighted in our introduction, real-world scenarios often involve a complex interplay between modalities and concurrent events.

Current audio-visual TAL models employ two-stage late fusion strategies, where audio and visual modalities interact only at the final classification stage. This approach reduces their ability to model fine-grained temporal dependencies, making them less effective for complex event localization. Some recent approaches incorporate cross-attention mechanisms~\cite{ramazanova2023owl,xue2021audio, wu2019dual,liu2022dense} but operate at fixed temporal scales and lack dynamic fusion control. 

Furthermore, contrastive learning techniques~\cite{hu2022mix, wang2024internvideo2} often focus on instance-level alignment while neglecting intra-video relationships, such as temporal coherence and cross-event correlations, which are crucial for distinguishing similar events occurring at different times. This limitation affects their ability to handle multi-scale variations and modality-specific characteristics.
Our DEL framework explicitly models cross-modal dependencies to address these challenges while preserving fine-grained temporal structure through an adaptive attention mechanism and a path aggregation network. Moreover, our dual contrastive learning strategy, incorporating inter-sample and intra-sample contrastive loss, refines feature discrimination within a single video, addressing the limitations of existing methods.

\section{Method Overview}
\label{sec:method}
\cref{fig:del_model} provides an overview of our framework. Given the untrimmed audio-visual input, the dense action localization task aims to precisely detect the temporal boundaries of action instances, requiring the refinement of their semantic representations. To begin with, we tokenize the audio and visual data using existing feature extraction networks. Our method then aligns and fuses audio-visual features for precise event localization through three key modules: (1) We employ an adaptive attention mechanism for cross-modal alignment, which dynamically aligns the audio and visual representations. (2) Next, score-based contrastive learning helps to dynamically select inter and intra mode hard-negative samples to improve feature discrimination in the training process. (3) Finally, our path aggregation network for multi-scale feature fusion facilitates multi-scale feature extraction by generating multiple temporal representations and aggregating them to enhance temporal modeling.

\subsection{TAL Problem Formulation}
We formulate dense audio-visual event localization as a joint event classification and boundary estimation problem.  Given a video with audio-visual data, we segment it into $T$ paired segments, represented as
\begin{equation}
\mathbb{S} = \{(\mathbf{V_t}, \mathbf{A_t})\}_{t=1}^T, 
\end{equation}
where $\mathbf{V_t}$ and $\mathbf{A_t}$ denote the visual and audio features at time $t$.  Note that the number of segments, $T$, varies across videos. 
The ground-truth annotation is defined as
\begin{equation}
\mathcal{G} = \{g_n = (\tau_{start, n}, \tau_{end, n}, \lambda_n)\}_{n=1}^N,
\end{equation}
where $\tau_{start, n}$ and $\tau_{end, n}$ indicate the event boundaries and $\lambda_n \in \Lambda$ represents the the event class from a defined set of $|\Lambda|$ categories and $N$ is the total number of events in the video.

During inference, our model predicts localized events 
\begin{equation}
\hat{\mathbb{S}} = \{\hat{s}_t = (\delta_{start, t}, \delta_{end, t}, q(y_t))\}_{t=1}^T
\end{equation}
where, $q(y_t) \in [0, 1]^{|\Lambda|}$ is the event classification probability, and $\delta_{start, t}$ and $\delta_{end, t}$ represent estimated temporal offsets of the current time $t$ for event boundaries. The final predictions are:
\begin{equation}
\begin{aligned}
\hat{\lambda}_t &= \arg\max_{\lambda \in \Lambda} q(\lambda_t), \\
\hat{\tau}_{start, t} &= t - \delta_{start, t}, \\
\hat{\tau}_{end, t} &= t + \delta_{end, t}.
\end{aligned}
\end{equation}
This results in the fine-grained localization and classification of overlapping audio-visual events.

\subsection{Adaptive Attention for Cross-Modal Alignment}
To address misalignment between audio and visual data in the videos, we propose an adaptive attention mechanism that dynamically adjusts feature importance, ensuring effective synchronization of corresponding audio and visual signals and enhancing cross-modal coherence. To compute this adaptive attention,  an attention mask $\mathbf{M}$, is initialized with the sizing, $ \mathbf{M} \in \mathbb{R}^{(L_v+L_a) \times (L_v+L_a)} $, where \( L_v \) and \( L_a \) denote the lengths of the video and audio features from the sequences, respectively. Then, given a concatenated input $\mathbf{X} =[\mathbf{V}|\mathbf{A}] \in \mathbb{R}^{(L_v + L_a) \times d}$ where X contains both video and audio features, we construct the query ($\mathbf{Q}$) and key ($\mathbf{K}$) matrices via linear transformations:
\begin{align}
\mathbf{Q} &= \mathbf{X} \mathbf{W}_Q, \quad \mathbf{K} = \mathbf{X} \mathbf{W}_K, \quad \mathbf{V} = \mathbf{X} \mathbf{W}_V
\end{align}
where $\mathbf{W}_Q, \mathbf{W}_K, \mathbf{W}_V \in \mathbb{R}^{d \times d}$ are learnable projection matrices.
Next, we compute the adaptive attention as:

\begin{align}
% Q &= X W_Q, \quad K = X W_K, \quad V = X W_V, \\
\text{aat}_{i,j} &= \frac{m_{i,j} \exp(\mathbf{Q}_i \mathbf{K}^T_j / \sqrt{d})}{\sum_k m_{i,k} \exp(\mathbf{Q}_i \mathbf{K}^T_k / \sqrt{d})} 
% \\
% Z &= X + D V W_O,
\end{align}
where \( i, j \in [1, L_v + L_a] \) represent the indices of the adaptive attention matrix, $\mathbf{AAT}$.
% and \( W_Q, W_K, W_V, W_O \in \mathbb{R}^{M \times M} \) are the linear projection matrices used to generate the query, key, value, and output vectors. We also employ multi-head attention~\cite{vaswani2017attention} to enhance the model’s capacity to capture complex relationships. This ensures that audio and visual signals corresponding to the same event are weighted properly, improving cross-modal coherence.
 The mask $\mathbf{M}$ aims to emphasize and learn feature interactions corresponding to the same event across both intra and inter modality. For intra-modality, we apply a standard global attention mechanism, allowing features within the same modality to attend to each other by setting the corresponding entries in the mask to 1. For cross-modality, only entries corresponding to the same temporal segment are assigned a value of 1, ensuring focused interactions between aligned video and audio features. The attention mask is then applied to the adaptive attention matrix from the self-attention mechanism, guiding modality-specific and cross-modal feature interactions.

\subsection{Score-based Contrastive Learning}
We adopt a score-based contrastive learning objective to improve the relationships between the audio-visual features within a video at training time. Unlike standard contrastive learning methods, which rely on predefined sampling heuristics, our method dynamically selects contrastive pairs using a learned score function that ranks the audio-visual features based on contextual similarity and temporal alignment. We use both inter-sample and intra-sample objectives. The inter-sample strengthens relationships between similar input video and audio pairs in a batch, while intra-sample learning captures fine-grained temporal relationships within the same sample in a single data modality.

\noindent \textbf{Inter-Sample Contrastive Learning.}
Given a batch of sample pairs, we employ an inter-sample contrastive loss that maximizes the cosine similarity between the [$\mathit{CLS}_V$] and [$\mathit{CLS}_A$] tokens of matched video-audio pairs while minimizing the similarity between tokens from mismatched pairs. 
The [$\mathit{CLS}_V$] and [$\mathit{CLS}_A$] tokens are global representations of the video and audio features, respectively. They are extracted from the adaptive attention layer of their respective modality and designed to capture high-level semantic information. 
%By aligning [$CLS_V$] and [$CLS_A$] tokens from matched video-audio pairs, we enforce cross-modal consistency, facilitating robust joint feature learning. 
The inter-sample contrastive loss is defined as:

\begin{align}
\mathcal{L}_{\text{inter}} &= \mathbb{E}_{\mathit{z} \sim [\mathit{CLS}_V]_{\mathit{j}}, \mathit{z}^+ \sim [\mathit{CLS}_A]_{\mathit{j}}, \mathit{z}^- \sim I_{\mathit{k} \neq \mathit{j}} [\mathit{CLS}_A]_{\mathit{k}}} \ell(\mathit{z}, \mathit{z}^+, \mathit{z}^-) \notag \\
& + \mathbb{E}_{\mathit{z} \sim [{CLS}_A]_{j}, z^+ \sim [\mathit{CLS}_V]_{j}, z^- \sim I_{k \neq j} [\mathit{CLS}_V]_{\mathit{k}}} \ell(\mathit{z}, \mathit{z}^+, z^-)
\label{eq:l_inter}
\end{align}

where \( \ell(z, z^+, z^-) \) represents the standard contrastive loss~\cite{he2020momentum}, defined by the following equation, and \( \tau \) is a learnable temperature parameter.

\begin{equation}
\resizebox{0.99\hsize}{!}{$
\ell(z, z^+, z^-) = 
-\log \left( \frac{\exp(z^T \cdot z^+ / \tau)}
{\exp(z^T \cdot z^+ / \tau) + k \exp(z^T \cdot z^-_k / \tau)} \right)$}
\label{eq:scl}
\end{equation}

\noindent \textbf{Intra-Sample Contrastive Learning with Score-Based Pair Selection.} 
\begin{figure}[tb]
  \centering
  \includegraphics[width=1.01\linewidth]{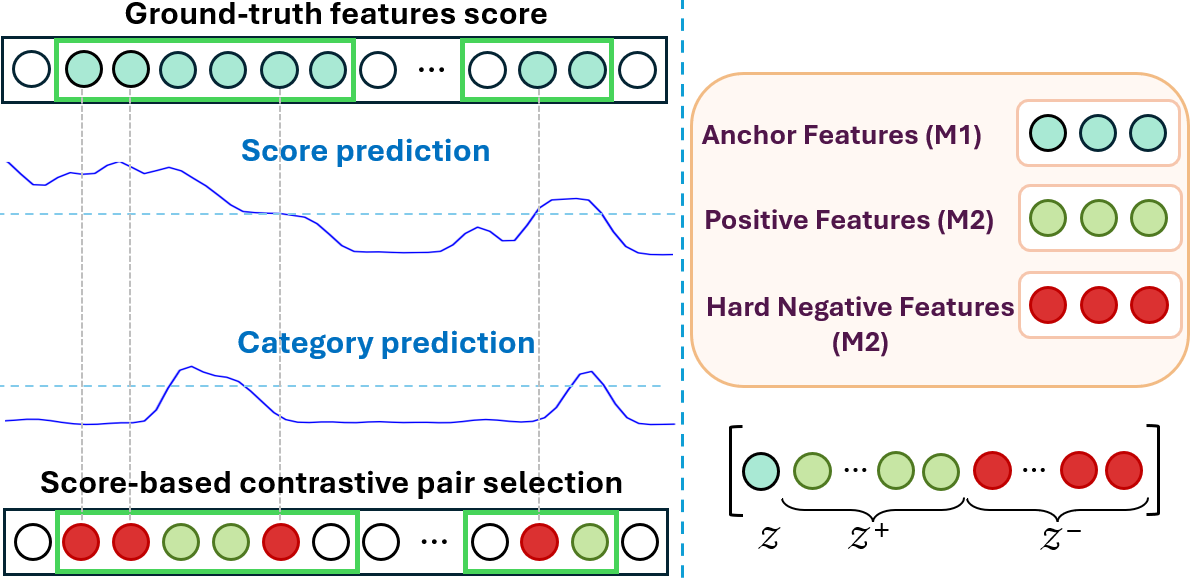}
  % \vspace{3pt}
  \caption{Score-based contrastive pair selection for identifying positive and hard-negative samples for each anchor within the event segment of a single modality. M1 represents the modality chosen as the anchor, while the goal is to select corresponding positive and hard-negative features from another modality represented as M2. Token-level predictions (score $s_t$ and category $c_t$) provide early supervision that refines latent features. This early processing guides the contrastive selection of hard-negative samples.}
  \label{fig:intra}
  % \vspace{-0.5 cm}
\end{figure}

As shown in \cref{fig:intra}, unlike inter-sample contrastive learning, score-based intra-sample contrastive learning enforces alignment within the same video, ensuring the model learns fine-grained feature distinctions. 
To achieve this, we introduce two token-level prediction mechanisms early in the network: a binary score function $s_t$ and an event category predictor $c_t$.
We define the score function $s_t$ as:
\begin{equation}
  s_t = \begin{cases}
            1 & \text{if $t$ belongs to the ground-truth event segment}\\
            0 & \text{otherwise}
         \end{cases}
\end{equation}

This function is trained as a binary classifier using ground-truth labels derived from event annotations. Features within an annotated event segment are assigned a score of 1, while those outside are assigned zero via a binary cross-entropy loss.

In parallel, the model predicts an event category $c_t$ for each feature using a cross-entropy loss over a predefined set of event classes. By jointly training $s_t$ and $c_t$ at the token level, the model learns both to identify the presence of an event and to classify its action, thereby refining its latent representations before final timestamp-wise decisions. 

These refined token-level predictions are key to selecting meaningful samples for the intra-contrastive loss.
Positive samples are identified as features where the model accurately predicts a score of 1 (i.e., $s_t$=1) and correctly classifies the event category ($c_t$), ensuring alignment with the ground-truth event segments in both video and audio modalities. Hard-negative samples are features that receive a high confidence score yet are assigned an incorrect event category, challenging the model to discern subtle differences.

The process results in the identification of positive visual features \(I_{\mathit{PV}} \), hard-negative visual features \(I_{\mathit{HNV}} \), positive audio features \(I_{\mathit{PA}} \), and hard-negative audio features \(I_{\mathit{HNA}} \). These samples are then utilized in an intra-sample contrastive loss function, following a similar contrastive equation defined in \cref{eq:scl}, to enhance the model's discriminative capabilities across modalities. The intra-contrastive loss is then formulated as:
\begin{align}
\mathcal{L}_{\text{intra}} &= \mathbb{E}_{\mathit{z} \sim I_{\mathit{PV}}, \mathit{z}^+ \sim I_{\mathit{PA}}, \mathit{z}^- \sim I_{\mathit{HNA}}} \ell(\mathit{z}, \mathit{z}^+, \mathit{z}^-) \notag \\
&\quad + \mathbb{E}_{z \sim I_{\mathit{PA}}, \mathit{z}^+ \sim I_{\mathit{PV}}, z^- \sim I_{\mathit{HNV}}} \ell(\mathit{z}, \mathit{z}^+, \mathit{z}^-)
\end{align}

This score-guided selection process ensures that positive samples are strongly aligned, while hard negatives provide challenging learning signals for better feature separation. Integrating hard-negative mining within intra-contrastive learning strengthens the model’s ability to differentiate between visually and aurally similar events, making it highly effective for dense event detection in untrimmed videos.

\subsection{Path Aggregation Network for Multi-Scale Feature Fusion}
%We employ a path aggregation network that enhances feature fusion across multiple temporal scales to effectively capture events of different durations. Rather than simply concatenating features, our path aggregation network ensures that information is propagated hierarchically, preserving both short-term details and long-term contextual dependencies.

Inspired by feature pyramid networks~\cite{yu2022mm,cheng2024yolo,geng2023dense}, we construct a multi-scale representation where lower levels focus on fine-grained temporal details while higher levels encode broader contextual information. 
%To further refine the interaction between audio and visual features, 
We introduce two key components: (I) \textbf{Modality-Guided Adapters}, which enriches visual features with relevant audio cues and vice versa, ensuring better cross-modal representation. These are followed by the (II) \textbf{Adaptive Pooling Module} that dynamically adjusts feature importance across different scales, emphasizing crucial temporal cues. 

In particular, the architecture processes audio features \((A_{l_1}, \dots, A_{l_n})\) and visual features \((V_{l_1}, \dots, V_{l_n})\) through top-down and bottom-up pathways to retain lower-level fine-grained details while aggregating higher-level contextual information. The modality-guided adapters employ a max-sigmoid attention mechanism to integrate audio cues into visual features,
\begin{equation}
V_l' = V_l \cdot \sigma\!\Bigl(\max_{j}\bigl(V_l A_j^\top\bigr)\Bigr)^\top,
\end{equation}
and similarly merge visual cues into audio embeddings,
\begin{equation}
A_l' = A_l \cdot \sigma\!\Bigl(\max_{k}\bigl(A_l V_k^\top\bigr)\Bigr)^\top,
\end{equation}
where \(\sigma\) denotes the sigmoid activation. The updated feature maps \(V_l'\) and \(A_l'\) are then combined with outputs from adjacent scales, propagating essential information throughout the pyramid. Finally, the adaptive pooling module aggregates multi-scale representations into compact tokens, \(\tilde{V}\) and \(\tilde{A}\), and refines each modality through multi-head attention,
\begin{equation}
\begin{aligned}
V' = V + \mathrm{MHA}(V, \tilde{A}, \tilde{A}) \\ 
\quad
A' = A + \mathrm{MHA}(A, \tilde{V}, \tilde{V})
\end{aligned}
\end{equation}
where \(\mathrm{MHA}\) denotes multi-head attention. This hierarchical design systematically fuses audio and visual cues at varying temporal scales, leading to more robust event localization and classification.

% To enhance the interplay between visual and audio features, our network introduces two key components: (I) a \textbf{Audio \& Visual-Guided Adapters} and (II) an \textbf{Adaptive Pooling Module}. The Audio-Guided Adapter enriches visual features with acoustic context, while the Visual-Guided Adapter enhances audio representations with visual semantics. These are followed by the Adaptive Pooling Module that refines cross-modal integration by dynamically adjusting feature importance across different scales, ensuring optimal weighting of short- and long-duration event cues. 

\subsection{Overall Objective Function}
The fused features are then passed through classification and regression heads for event category prediction and temporal boundary refinement. The final learning objective of DEL is a combination of the contrastive inter and intra losses, $\mathcal{L}_{inter}$ and $\mathcal{L}_{intra}$, and the score cross entropy loss $\mathcal{L}_{score}$. 
Additionally, the classification head is trained using a cross-entropy loss, $\mathcal{L}_{\text{cls}}$, which ensures accurate event categorization, while the regression head is optimized with a smooth L1 loss, $\mathcal{L}_{\text{reg}}$, to refine the temporal boundaries of each detected event: 

\begin{align}
  \mathcal{L}_\mathit{DEL} =
  \lambda_{1} \mathcal{L}_\mathit{inter} &+ \lambda_{2} \mathcal{L}_\mathit{intra}+\lambda_{3} \mathcal{L}_\mathit{score} \notag \\
  &+\lambda_{4} \mathcal{L}_\mathit{cls}+\lambda_{5} \mathcal{L}_\mathit{reg}  
  \label{eq:loss}
\end{align}
    
$\lambda$ values adjust the weighting between each loss term. The values are set to balance loss terms, ensuring that no single term dominates optimization while maintaining model stability and convergence.

\begin{table}[tbp]
\centering

   % \centering
  \scalebox{0.76}{
\begin{tabular}{l c c c c c c c}
\toprule
\multicolumn{1}{c}{Method}                        &  & $0.3$ & $0.4$ & $0.5$ & $0.6$ & $0.7$ & Avg\\ \hline

%TAL-MR~\cite{zhao2020bottom}  &  & $53.9$ & $50.7$ & $45.4$ & $38.0$ & $28.5$ & $43.4$ \\
%P-GCN~\cite{zeng2019graph}  &  & $63.6$ & $57.8$ & $49.1$ & - & - & - \\
MUSES~\cite{liu2021multi}  &  & $68.9$ & $64.0$ & $56.9$ & $46.3$ & $31.0$ & - \\
ContextLoc~\cite{zhu2021enriching}  &  & $68.3$ & $63.8$ & $54.3$ & $41.8$ & $26.2$ & $50.9$  \\    
%RTD-Net~\cite{tan2021relaxed}  &  & $68.3$ & $62.3$ & $51.9$ & $38.8$ & $23.7$ & $49.0$\\
$A^2$Net~\cite{yang2020revisiting}  &  & $58.6$ & $54.1$ & $45.5$ & $32.5$ & $17.2$ & $41.6$\\
PBRNet~\cite{liu2020progressive}  &  & $58.5$ & $54.6$ & $51.3$ & $41.8$ & $29.5$ & - \\
AFSD~\cite{lin2021learning}  &  & $67.3$ & $62.4$ & $55.5$ & $43.7$ & $31.1$ & $52.0$\\
TadTR~\cite{liu2022end}  &  & $62.4$ & $57.4$ & $49.2$ & $37.8$ & $26.3$ & $46.6$ \\
Actionformer~\cite{zhang2022actionformer}  &  & $82.1$ & $77.8$ & $71.0$ & $59.4$ & $43.9$ & $67.9$\\
ASL~\cite{shao2023action}  &  & $83.1$ & $79.0$ & $71.7$ & $59.7$ & $45.8$ & $66.8$ \\
TMaxer+MRAVFF~\cite{fish2023multi}  &  & $82.2$ & $78.2$ & $71.5$ & $59.9$ & $45.3$ & $67.4$ \\
TriDet~\cite{shi2023tridet}  &  & $\mathbf{83.6}$ & $\mathbf{80.1}$ & $\mathbf{72.9}$ & $62.4$ & $47.4$ & $69.3$\\ \midrule
\textbf{DEL} &  & $81.0$ & $78.0$ & $71.8$ & $\mathbf{68.4}$ & $\mathbf{60.5}$ & $\mathbf{71.9}$\\
  \bottomrule
\end{tabular}
}
\caption{\textbf{Performance comparison on THUMOS14} We report mAP across multiple tIoU thresholds and compute the average mAP. Our method outperforms previous approaches on THUMOS14 with the same feature extraction. }
\label{tab:thumos}
\end{table}

\section{Experiments and Results}

\noindent \textbf{Dataset and Metrics} We conduct comprehensive evaluations on four key benchmarks: THUMOS14~\cite{idrees2017thumos}, ActivityNet-1.3~\cite{caba2015activitynet}, EPIC-Kitchens-100~\cite{damen2022rescaling}, and UnAV-100~\cite{geng2023dense}. We follow the standard evaluation protocol, measuring performance using mean Average Precision (mAP) across multiple temporal Intersection over Union (tIoU) thresholds. To ensure robustness and reliability, we report averaged performance from five independent training runs to mitigate the impact of initialization.

\noindent \textbf{Feature Encoder.} We follow existing works, and for visual feature extraction, we used I3D~\cite{carreira2017quo}, a two-stream inflated 3D convolutional network pretrained on Kinetics-400, for THUMOS14, ActivityNet, and UnAV-100, leveraging its effectiveness in capturing spatio-temporal dynamics. Like other works~\cite{zhang2022actionformer,shao2023action,shi2023tridet} for EPIC-Kitchens-100, where fine-grained action recognition is crucial, we employed the SlowFast network\cite{feichtenhofer2019slowfast}, pretrained on EPIC-Kitchens. Across all experiments, audio features were extracted using the VGGish model, pretrained on AudioSet~\cite{gemmeke2017audio}, ensuring consistent and high-quality audio representations. 
% Audio and video features are projected into a shared embedding space via a fully connected layer. 
% \begin{equation}
% V = \left\{ v_t \right\}_{t=1}^{T} \in \mathbb{R}^{T \times D}, 
% \quad
% A= \left\{ a_t \right\}_{t=1}^{T} \in \mathbb{R}^{T \times D}.
% \end{equation}
%  where D is the dimension of the embedding space.

\subsection{Quantitative Results}

We first show the performance of our method across the range of datasets. \cref{tab:thumos} shows how on THUMOS14, our method (DEL) achieves a state-of-the-art average mAP of 71.9\% (mAP@[0.3:0.1:0.7]), surpassing prior methods such as TriDet by +2.6\%. Notably, DEL demonstrates superior performance at higher IoU thresholds, achieving 68.4\% at 0.6 and 60.5\% at 0.7, significantly outperforming existing approaches. This highlights DEL’s ability to localize actions precisely.

\begin{table}[tbp]
\centering
  \scalebox{0.85}{
      \raggedleft
\begin{tabular}{c c c c c c c}
\toprule
\multicolumn{1}{c}{Method} &  & 0.5 & 0.75       & 0.95      & Avg      \\ \hline

%TAL-MR~\cite{zhao2020bottom}   &  & 43.5 & 33.9 & 9.2 & 30.2  \\
%P-GCN~\cite{zeng2019graph}   &  & 48.3 & 33.2 & 3.3 & 31.1  \\
MUSES~\cite{liu2021multi}  &  & 50.0 & 35.0 & 6.6 & 34.0 \\
ContextLoc~\cite{zhu2021enriching}  &   & 56.0 & 35.2 & 3.6 & 34.2  \\
VSGN~\cite{zhao2021video}   & & 52.3 & 35.2 & 8.3 & 34.7 \\
%RTD-Net~\cite{tan2021relaxed}   &  & 47.2 & 30.7 & 8.6 & 30.8  \\
$A^2$Net~\cite{yang2020revisiting}   &  & 43.6 & 28.7 & 3.7 & 27.8 \\
PBRNet~\cite{liu2020progressive}   & & 54.0 & 35.0 & 9.0 & 35.0 \\
AFSD~\cite{lin2021learning}   &  & 52.4 & 35.3 & 6.5 & 34.4 \\
TadTR~\cite{liu2022end}   & & 49.1 & 32.6 & 8.5 & 32.3  \\
Actionformer~\cite{zhang2022actionformer}  &   & 53.5 & 36.2 & 8.2 & 35.6 \\
ASL~\cite{shao2023action}   &  & 54.1 & 37.4 & 8.0 & 36.2 \\
TriDet~\cite{shi2023tridet}   &  & 54.7 & 38.0 & 8.4 & 36.8 \\ \midrule
\textbf{DEL}  &  & \textbf{56.9} & \textbf{42.5} & \textbf{14.7} & \textbf{38.0} \\
  \bottomrule
\end{tabular}
}

\caption{\textbf{Performance evaluation on ActivityNet 1.3.} We present mAP and average mAP results across various tIoU thresholds. Our approach surpasses previous
methods with the same feature extraction.}
\label{tab:anet}
\end{table}

For ActivityNet-1.3, \cref{tab:anet}, DEL outperforms recent works with a +1.2\% gain in average mAP, consistently improving across IoU thresholds. DEL performs better than prior methods, demonstrating its effectiveness in handling large-scale datasets with diverse and long-duration activities. The model’s ability to refine event boundaries and integrate multi-scale temporal cues contributes to this improvement.

\begin{table}[!htbp]
\centering
    \resizebox{8.5cm}{!}{
    \raggedleft
  \centering
  \scalebox{10}{
  \begin{tabular}{c c c c c c c c}
    \toprule
    Task & Method & 0.1 & 0.2 & 0.3 & 0.4 & 0.5 & Avg\\
    \midrule
      & BMN~\cite{lin2019bmn} & 10.8 & 8.8 & 8.4 & 7.1 & 5.6 & 8.4\\
      & G-TAD~\cite{xu2020g} & 12.1 & 11.0 & 9.4 & 8.1 & 6.5 & 9.4\\
     Verb & ActionFormer~\cite{zhang2022actionformer} & 26.6 & 25.4 & 24.2 & 22.3 & 19.1 & 23.5\\
    & ASL~\cite{shao2023action} & 27.9 & - & 25.5 & - & 19.8 & 24.6\\
    & ActionFormer + MRAV-FF~\cite{fish2023multi}&  27.6 & 26.8 & 25.3 & 23.4 & 19.8 & 24.6\\
      & TriDet~\cite{shi2023tridet} & 28.6 & 27.4 & 26.1 & 24.2 & 20.8 & 25.4\\ \midrule
      & \textbf{DEL} & \textbf{32.2} & \textbf{29.9} & \textbf{27.8} & \textbf{25.1} & \textbf{20.8} & \textbf{27.1} \\
  % \bottomrule
 
    %\toprule
     \midrule
      & BMN~\cite{lin2019bmn} & 10.3 & 8.3 & 6.2 & 4.5 & 3.4 & 6.5\\
      & G-TAD~\cite{xu2020g} & 11.0 & 10.0 & 8.6 & 7.0 & 5.4 & 8.4\\
     Noun & ActionFormer~\cite{zhang2022actionformer} & 25.2 & 24.1 & 22.7 & 20.5 & 17.0 & 21.9\\
      & ASL~\cite{shao2023action} & 26.0 & - & 23.4 & - & 17.7 & 22.6\\
        & ActionFormer + MRAV-FF~\cite{fish2023multi}&  26.4 & 25.4 & 23.6 & 21.2 & 17.4& 22.8\\
      & TriDet~\cite{shi2023tridet} & 27.4 & 26.3 & 24.6 & 22.2 & 18.3 & 23.8\\ \midrule
      & \textbf{DEL} & \textbf{29.5} & \textbf{28.4} & \textbf{26.2} & \textbf{22.9} & \textbf{19.3} & \textbf{25.2} \\
  \bottomrule
  \end{tabular}
  
  }

    }
      \caption{\textbf{Performance on the EPIC-Kitchens-100 validation set} across multiple tIoU thresholds, with average mAP reported. Our method outperforms all baselines by a significant margin using the same feature extraction.}
    \label{tab:epic}
\end{table}

On EPIC-Kitchens-100, \cref{tab:epic}, a challenging dataset focused on fine-grained cooking activities, DEL achieves 27.1\% and 25.2\% average mAP for the verb and noun tasks, respectively. Compared to THUMOS14 and ActivityNet, EPIC-Kitchens presents a more complex setting with lower IoU thresholds (0.1–0.5), requiring finer temporal precision and handling of subtle action variations. DEL’s performance gain in this setting demonstrates its strength in distinguishing short, overlapping interactions through effective cross-modal alignment.

\begin{table}[tb]
\centering

    \resizebox{8.3cm}{!}{
    \raggedleft
  \centering
  \scalebox{10}{
  \begin{tabular}{c c c c c c c}
    \toprule
     Method & 0.5 & 0.6 & 0.7 & 0.8 & 0.9 & Avg\\
    \midrule
      VSGN~\cite{zhao2021video} & 24.5 & 20.2 & 15.9 & 11.4 & 6.8 & 24.1\\
      TadTR~\cite{liu2022end} & 30.4 & 27.1 & 23.3 & 19.4 & 14.3 & 29.4\\
     ActionFormer~\cite{zhang2022actionformer} & 43.5 & 39.4 & 33.4 & 27.3 & 17.9 & 42.2\\
      UnAV~\cite{geng2023dense} & 50.6 & 45.8 & 39.8 & 32.4 & 21.1 & 47.8\\ \midrule
      \textbf{DEL} & \textbf{53.4} & \textbf{48.1} & \textbf{42.6} & \textbf{35.6} & \textbf{26.9} & \textbf{51.1} \\
  % \bottomrule
  \bottomrule
  \end{tabular}
  }
    }
    
  \caption{\textbf{Performance on the UnAV-100 test set}, showcasing our method's significant improvement over all baselines using the same feature extraction. We report mAP and average mAP at various tIoU thresholds.}
  \label{tab:unav}
\end{table}

Finally, on UnAV-100, \cref{tab:unav}, a dataset featuring complex multi-event scenarios and significant audio-visual overlap, DEL achieves a state-of-the-art average mAP of 51.1\%, outperforming previous methods for 3.3\%. The model’s ability to capture cross-modal dependencies and adaptively fuse features across multiple temporal scales enables the robust localization of overlapping and concurrent events.

Across all benchmarks, DEL consistently improves as the IoU threshold increases, demonstrating its ability to refine temporal boundaries and focus on more confident, well-localized events.
This characteristic ensures higher precision and fewer false positives, making DEL a robust choice for real-world applications requiring fine-grained audio-visual event localization.

\subsection{Ablation Experiments}
To explore the model's performance in more detail, we ran an in-depth analysis on UnAV-100. This dataset was selected due to its complex audiovisual interactions, overlapping events, and untrimmed video format. Our study reveals that removing key modules leads to a significant performance drop, whereas multi-scale fusion and feature quality enhancement improve localization accuracy.

\noindent\textbf{Component Ablation.} 
\begin{table}[htb!]
\centering
    \resizebox{7.5cm}{!}{
    \raggedleft
  \centering

  \begin{tabular}{c c c | c c c c c c c}
    \toprule
     AAC & SCL & PAN & 0.5 & 0.6 & 0.7 & 0.8 & 0.9 & Avg\\
    \midrule
     $\times$  &$\checkmark$ & $\checkmark$& 51.1 & 45.7 & 41.0 & 34.5 & 25.8 & 49.6 \\
      $\checkmark$ &$\times$  & $\checkmark$ & 51.5 & 44.7 & 38.7 & 33.3 & 25.8 & 49.7 \\
    $\checkmark$ &$\checkmark$ & $\times$ & 51.3 & 45.0 & 39.4 & 33.8 & 25.3 & 49.5\\
          \rowcolor{gray!20!}$\checkmark$ &$\checkmark$ & $\checkmark$ & 53.4 & 48.1 & 42.6 & 35.6 & 26.9& 51.1 \\

  % \bottomrule
  \bottomrule
  \end{tabular}
  }
 \caption{\textbf{Component-wise ablation study}, evaluating the individual contributions of our proposed Adaptive Attention for Cross-Modal Alignment (AAC), Score-Based Contrastive Learning (SCL), and Path Aggregation Network for Multi-Scale Feature Fusion (PAN) modules.}
\label{tab:comp}
\end{table}
\cref{tab:comp} presents the results of removing key components, namely the adaptive attention mechanism, score-based contrastive learning, and path aggregation module. The analysis highlights the relative performance gains achieved by each component, demonstrating the effectiveness of DEL in improving generalizability and refining event localization.

\noindent\textbf{Pyramid Levels.}
\begin{table}[htb!]
\centering
    % \centering
    \resizebox{6.5cm}{!}{
    \raggedleft
  \centering

  \begin{tabular}{c | c c c c c c c}
    \toprule
      $L$ & 0.5 & 0.6 & 0.7 & 0.8 & 0.9 & Avg\\
    \midrule
      1 & 47.5 & 42.7 & 37.3 & 30.6 & 22.0 & 45.5 \\
      2 & 48.5 & 43.2 & 37.7 & 31.2 & 23.0 & 46.4 \\
     4 & 48.4 & 43.5 & 38.2 & 32.0 & 23.5 & 47.2\\
          \rowcolor{gray!20!} 6 &  53.4 & 48.1 & 42.6 & 35.6 & 26.9& 51.1 \\
       7 & 51.0 & 45.9 & 40.1 & 33.9 & 24.6 & 49.0 \\
  % \bottomrule
  \bottomrule
  \end{tabular}
  }
 \caption{Ablation study on the design of the feature pyramid. $L$ shows the number of layers for both audio and video.}
\label{tab:levels}
\end{table}
The path aggregation module fuses features across multiple scales within our DEL framework. \cref{tab:levels} details the results of evaluating our model on the UnAV-100 dataset using varying levels for visual and audio. Our analysis reveals that utilizing six pyramid levels for both modalities yields the best performance.  These results suggest a critical balance; too few levels limit the capture of multi-scale context, while excessive levels introduce redundant or noisy information. \cref{tab:levels} reinforces that multi-scale audio-visual fusion and optimization of pyramid level design are critical for robust action localization performance.

\noindent\textbf{Enhancing Feature Quality.}
To achieve precise event localization, the model must learn high-quality feature representations that capture modality-specific and cross-modal information. Enhancing feature quality ensures better differentiation between similar events and improves robustness against temporal inconsistencies, ultimately leading to more accurate and reliable predictions.
Therefore, to explore the influence of the audio-visual feature extractors, we replace I3D with DINOv2~\cite{oquab2023dinov2} for video and MERT-v1~\cite{li2023mert} for audio. The results shown in \cref{tab:thumos-merd} demonstrate that by integrating DINOv2 for video and MERT for audio, the performance does improve slightly, making our proposed method suitable for use with multiple feature extractor models.

\begin{table}[htbp]
\centering

% \centering
  \scalebox{.85}{
      \raggedleft
\begin{tabular}{ c c c c c c c}
\toprule
      Features & 0.3 & 0.4 & 0.5 & 0.6 & 0.7 & Avg\\
 \midrule
\multicolumn{7}{c}{THUMOS14}  \\
\midrule
I3D+Vggish & 81.0 & 78.0 & 71.8 & 68.4& 60.5 & 71.9 \\
DINOv2+MERT & 81.5 & 79.1  & 73.1 & 70.8 & 64.6&  73.3  \\ \midrule
\multicolumn{7}{c}{UnAV-100}  \\
\midrule
I3D+Vggish& 53.4 & 48.1 & 42.6 & 35.6 & 26.9 & 51.1 \\
DINOv2+MERT & 55.0 & 49.7 & 44.1 & 37.4 & 28.4 & 52.7 \\
\bottomrule
\end{tabular}
}
\caption{ Evaluation on THUMOS14 and UnAV-100 incorporating DINOv2 for video features and MERTv1 for audio features.}
\label{tab:thumos-merd}
\end{table}

\noindent\textbf{Impact of Different Input Modalities}
A core contribution of this work is the effective fusion of audio and visual modalities within the DEL framework, enabling robust event localization in complex, untrimmed videos. As shown in \cref{tab:mod}, DEL achieves a significantly superior average mAP of 51.1\%  when leveraging audio and video inputs. This represents a substantial performance increase compared to configurations using only audio (40.6\%) or only video (37.9\%), demonstrating the critical and complementary role of audio-visual information for accurate event localization. This performance gain is consistently observed across a comprehensive range of IoU thresholds (0.5 to 0.9), further underscoring the robustness and effectiveness of DEL's multi-modal fusion strategy. These results validate the design choices in DEL and highlight the importance of multi-scale cross-modal perception and dependency modeling for advancing audio-visual scene understanding.
\begin{table}[htb!]
\centering
    \resizebox{8.4cm}{!}{
    \raggedleft
  \centering

  \begin{tabular}{ c c | c c c c c c c}
    \toprule
    Audio & Video & 0.5 & 0.6 & 0.7 & 0.8 & 0.9 & Avg\\
    \midrule
        $\checkmark$ &$\times$& 41.6  & 36.3 & 32.7 & 27.8 & 21.6 & 40.6 \\
      $\times$  & $\checkmark$ & 38.3 & 33.4 & 28.9 & 25.2 & 19.2 & 37.9 \\

          \rowcolor{gray!20!} $\checkmark$ & $\checkmark$ & 53.4 & 48.1 & 42.6 & 35.6 & 26.9& 51.1 \\

  % \bottomrule
  \bottomrule
  \end{tabular}
  }
 \caption{DEL performance with various modality combinations. Fusing audio and video yields the best results, emphasizing the importance of multi-modal input.}
\label{tab:mod}
\end{table}

\subsection{Qualitative Results.}

\begin{figure}[tb]
  \centering
\includegraphics[trim={3.01cm 6.1cm -4cm 3cm},clip,width=1.28\linewidth]{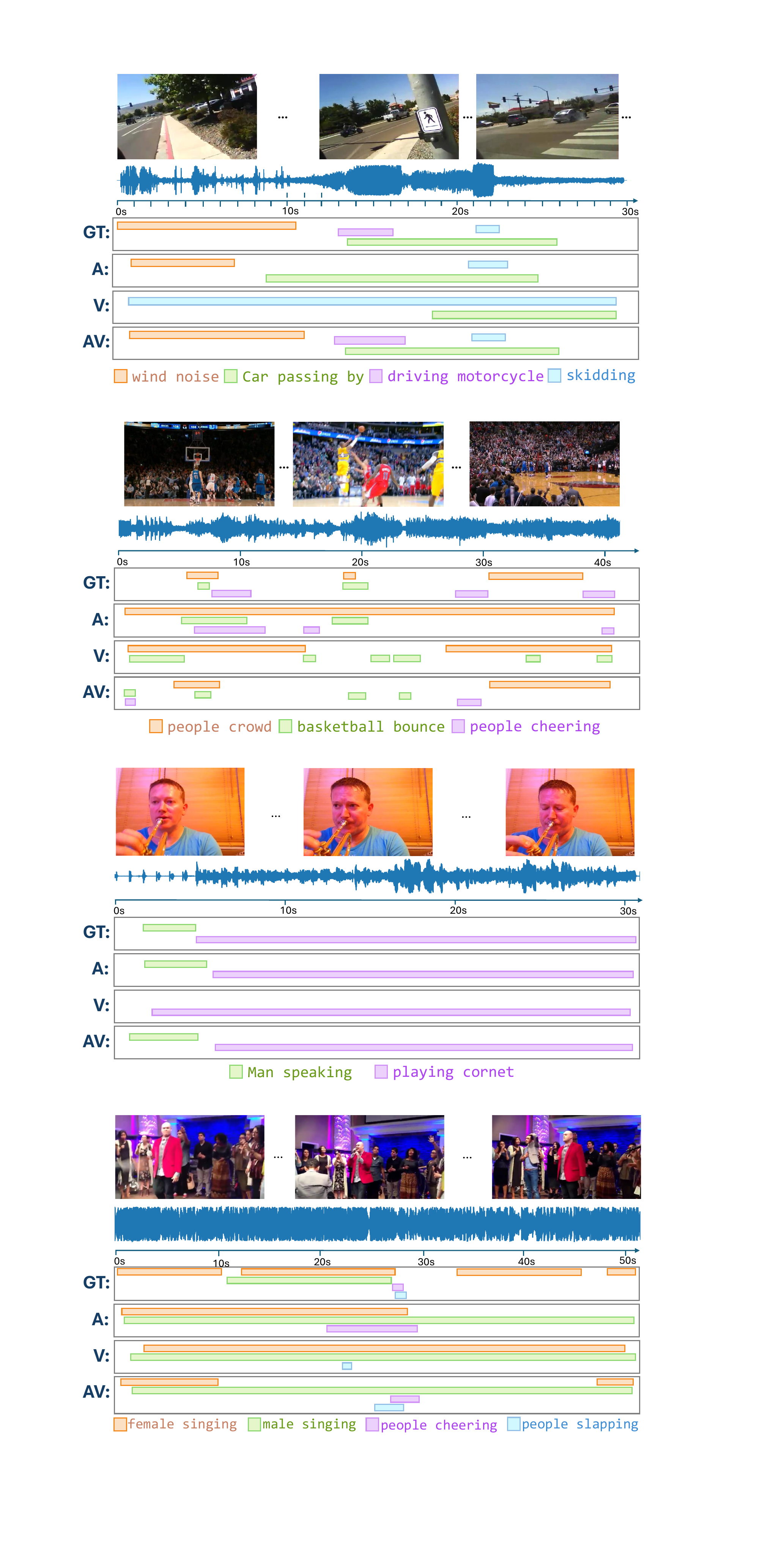}
  \vspace{-15pt}
  \caption{Qualitative results of our DEL framework for audio-visual event localization. We present ground-truth (GT) and event categories alongside predictions from audio-only (A), visual-only (V), and audio-visual (AV) models. The AV model (DEL) achieves more accurate event localization by effectively leveraging both modalities, while the unimodal models struggle with events that rely on cross-modal cues.}
  \label{fig:AV}
\end{figure}
In \cref{fig:AV}, we present qualitative results demonstrating the effectiveness of our DEL framework in comparison to unimodal baselines. For the first example, our model, leveraging audio and visual streams, accurately localizes events such as wind noise, cars passing by, driving a motorcycle, and skidding, even in scenarios with overlapping or short-duration occurrences. In contrast, the audio-only model struggles with visually-driven events like skidding, while the vision-only model fails to detect sound-based events like wind noise. Moreover, relying solely on audio leads to incorrect predictions, such as misclassifying the scene as engine knocking due to the absence of visual context. Similarly, the vision-only model, lacking critical audio cues, misinterprets the scene as auto racing from start to finish, based purely on visual perception. This shows the impact of audio in disambiguating visually similar activities. Distinguishing between "man speaking" and another potential sound source in the third scenario is only possible with audio input, as visual information alone is insufficient.

 Similarly, in the last example,  the scene is crowded, making it challenging to infer "people cheering" using only visual cues. The vision-only model struggles to recognize this category, while the audio modality provides crucial information for its correct identification. Additionally, detecting "people slapping" relies primarily on visual cues. These results highlight how integrating audio and visual streams leads to more accurate and robust event localization, particularly in complex multimodal scenarios.

These results highlight the complementary nature of audio and visual modalities for precise dense event localization, particularly in complex, real-world scenarios where events are often overlapping and context-dependent. 
% The audio-only model struggles with visual-dependent events like skidding, while the vision-only model fails to capture sound-based events like wind noise. 
% However, when using only the audio-based model, it incorrectly predicts the engine knocking class, as it heavily relies on sound cues without visual context. Similarly, the vision-only model struggles with audio-dependent events like wind noise and misclassifies the scene as auto racing from the beginning to almost the end of the video, as it solely depends on visual content.

% As visualized in Fig.~\ref{}, the qualitative results highlight the critical role of each module within our DEL framework. By comparing the full DEL model to ablated versions lacking the adaptive attention mechanism, score-based contrastive learning, or path aggregation, we observe a clear degradation in localization accuracy for single modality. In particular, scenarios with prolonged audio presence, such as auto racing, demonstrate the model's reliance on cross-modal information to accurately delineate event boundaries. These observations underscore the complementary nature of audio and visual modalities, validating the efficacy of our cross-modal fusion techniques in achieving robust and precise audio-visual event localization. Additional qualitative results are provided in the supplementary materials.
\section{Conclusion}
This paper introduced DEL, a dense audio-visual event localization framework for untrimmed videos. DEL effectively tackles the challenges of overlapping events and complex temporal dependencies by dynamically aligning audio and visual representations with an adaptive attention mechanism and a novel contrastive learning strategy. The vision-audio path aggregation network further strengthens cross-modal interactions, enabling deeper audio-visual feature integration at multiple temporal resolutions. With state-of-the-art performance on UnAV-100, THUMOS14, ActivityNet 1.3, and EPIC-Kitchens-100, DEL redefines fine-grained semantic action localization, paving the way for more sophisticated audio-visual scene understanding. DEL advances the field by bridging the gap between existing methodologies and real-world video analysis. It sparks new possibilities for future research in uncovering even richer insights from complex video content.

{
    \small
    \bibliographystyle{ieeenat_fullname}
    \bibliography{main}
}

% WARNING: do not forget to delete the supplementary pages from your submission 
\clearpage
\setcounter{page}{1}
\maketitlesupplementary

\section{Experimental Details}
\label{sec:rationale}

We present the implementation details, including the network architecture, training process, and inference strategy. Our code provides further information.

\subsection{Evaluation Metric: Mean Average Precision (mAP)}
In temporal action localization, mean Average Precision (mAP) is the most commonly used evaluation metric, with t-IoU = 0.5 as a standard comparison reference point.

Precision (P) measures the proportion of correctly detected action instances within a single class for a given video. Specifically, for class C, precision is defined as:
\begin{equation}
P = \frac{TP}{TP + FP} = \frac{\text{Number of correctly predicted proposals}}{\text{Total number of predicted proposals}}
\end{equation}

Since the test set contains multiple videos, Average Precision (AP) represents the mean precision for a specific class C across all test videos. Further, as the test set spans multiple action classes, the mean Average Precision (mAP) is computed as the mean of AP values across all classes:

\begin{equation}
mAP = \frac{\sum AP}{\text{Total number of classes}}
\end{equation}

In summary, under a given t-IoU threshold, Precision (P) quantifies the accuracy of detected action instances within a specific class in a single video, AP reflects the averaged precision across all classes in a video, and mAP generalizes this across all test videos and classes. Following standard evaluation protocols, most studies report mAP at multiple t-IoU thresholds to assess model performance comprehensively.

\begin{figure*}[t]
    \centering
    \includegraphics[trim={0cm 4cm 0cm 2cm}, height=5.1cm]{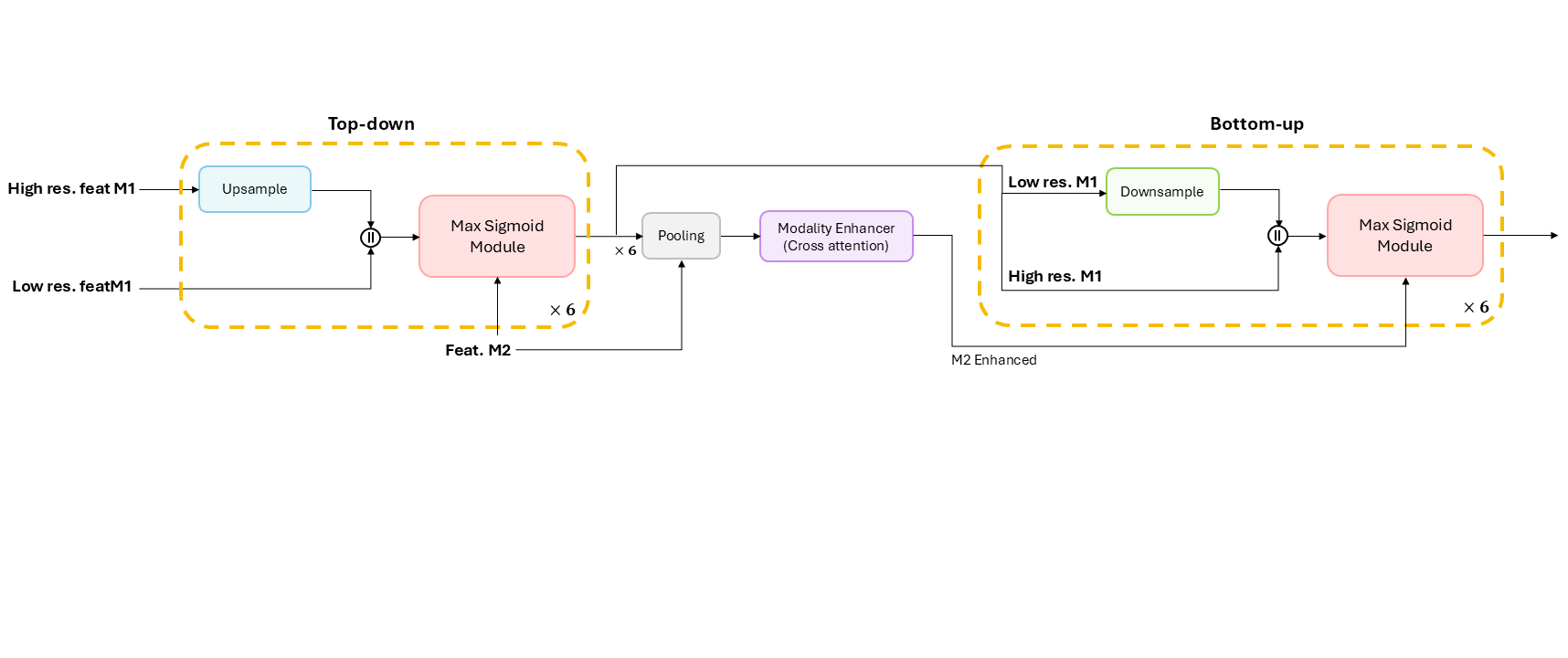}
      \vspace{-50pt}

    \caption{Illustration of the Path Aggregation Network for multi-scale feature fusion. The network employs a top-down and bottom-up pathway to fuse feature maps from different temporal scales, which helps capture fine-grained details and high-level semantics. A modality enhancer module refines cross-modal feature representations by applying cross-attention, ensuring robust integration of audio-visual data for accurate event localization. M1 indicates one modality (e.g., visual), and M2 represents the other (e.g., audio) 
 .$\mathbin{\|}$ represents the concatenation operation.  The details of the max sigmoid module are shown~\cref{fig:maxsig}. 
    %DEL efficiently localizes fine-grained and overlapping events in untrimmed videos, leveraging cross-modal dependencies for improved accuracy.
    }
    \label{fig:pan}
\end{figure*}

\subsection{Datasets} 

Our experimental setup differs across datasets due to variations in video resolution, frame rates, and feature types. Below, we provide specific details for THUMOS14, ActivityNet-1.3, EPIC-Kitchens-100, and UnAV100. THUMOS14 consists of 413 untrimmed videos from 20 action classes, utilizing 200 validation videos for training and 213 test videos for evaluation. ActivityNet v1.3, a large-scale dataset with over 20,000 videos, spans 200 action categories, containing 10,024 training videos and 4,926 validation videos. ActivityNet v1.2, a subset of v1.3, includes 100 action categories with 4,819 training and 2,383 validation videos. EPIC-Kitchens-100 contains 100 hours of cooking activities in kitchens. It has two tasks: noun localization (300 classes) and verb localization (97). While UnAV-100 contains 30,059 audiovisual events across 100 categories within 10,790 untrimmed videos, totaling over 126 video hours. Approximately 60\% of videos contain multiple audio-visual events, averaging 2.8 events per video, and around 25\% include concurrent events.

\begin{itemize}
    \item THUMOS14: For feature extraction on THUMOS14, we utilized a two-stream I3D model~\cite{carreira2017quo} pretrained on Kinetics, aligning with~\cite{zhang2022actionformer,zhao2020bottom}. Each input clip consisted of 16 consecutive frames, processed using a sliding window with a stride of 4. We extracted 1024-D features from the layer preceding the final fully connected layer and combined the two-stream outputs into a 2048-D representation as input to our model. Performance was assessed using mAP across tIoU thresholds in the range [0.3:0.1:0.7].

The model was trained for 50 epochs, starting with a 5-epoch linear warmup. The learning rate was initialized at 5e-4. To improve generalization, a cosine decay schedule with a weight decay of 1e-4 was applied.

\item \textbf{ActivityNet-1.3} We utilized a two-stream I3D model~\cite{carreira2017quo} for feature extraction, setting the sliding window stride to 16. Following previous studies~\cite{zhang2022actionformer, lin2018bsn, lin2019bmn}, features were downsampled to fixed lengths of 160 using linear interpolation for I3D features. Performance was measured using mAP@[0.5:0.05:0.95], along with the average mAP. The model was trained for 15 epochs with a 5-epoch linear warmup, using a learning rate of 5e-4, and a weight decay of 1e-4. For ActivityNet, we adopted a proposal generation strategy that groups all actions within a single category, followed by external classification scoring for recognition, a technique proven effective in prior single-stage temporal action localization approaches~\cite{lin2021learning}.

\item \textbf{EPIC-Kitchens-100}
We extracted features using a SlowFast network~\cite{feichtenhofer2019slowfast}, which was pretrained on EPIC-Kitchens-100 and provided by~\cite{damen2022rescaling}. The model processed 32-frame video segments with a stride of 16, generating 2304-dimensional feature embeddings. Training was conducted on the designated training split, while evaluation was performed on the validation set. 

For evaluation, we followed the mAP@[0.1:0.1:0.5] metric and reported the average mAP, maintaining consistency with~\cite{damen2022rescaling}. The model was trained for 30 epochs, using an initial learning rate of 5e-4 and a weight decay of 1e-4, ensuring stable convergence.

\item \textbf{UnAV100} For processing UnAV-100, we extract visual and audio features while ensuring proper temporal alignment. Video frames are sampled at 25 fps, and for each segment, we use a two-stream I3D~\cite{carreira2017quo} network to process 24 consecutive RGB and optical flow frames. A stride of 8 is applied during feature extraction, and the outputs from both streams are concatenated to form a 2048-dimensional visual representation. Simultaneously, VGGish~\cite{hershey2017cnn} extracts audio embeddings from 0.96-second segments with a stride of 0.32 seconds, ensuring synchronization with the visual features. Since video lengths vary, we standardize inputs by cropping or padding sequences to a fixed length of T = 224.

For training, we adopt the Adam optimizer and run the model for 40 epochs, incorporating a 5-epoch linear warmup at the start. The initial learning rate is set to 1e-3 and follows a cosine decay schedule for smoother optimization. We apply weight decay of 1e-4 to regularize training. Given that UnAV-100 focuses on temporal event localization in untrimmed videos, we evaluate model performance using mean Average Precision (mAP). Specifically, we compute mAP at tIoU thresholds from 0.5 to 0.9 (step size 0.1) and report the average mAP over the range 0.1 to 0.9, ensuring a robust assessment of localization accuracy.

\end{itemize}

\section{Path Aggregation Network for Multi-Scale Feature Fusion}

We propose a path aggregation network to ensure information is aggregated across multiple temporal resolutions, preserving short-term event cues and long-term contextual dependencies. This approach ensures greater feature consistency across different temporal resolutions, ultimately improving the regression of an event's location.

As illustrated in~\cref{fig:pan}, we employ top-down and bottom-up pathways to construct a feature pyramid. This design integrates multi-scale visual features (\(V_{l_1} \) to \(V_{l_n} \)) and multi-scale audio features (\(A_{l_1} \) to \(A_{l_n} \)) into a series of feature pyramids (\(P_{1} \) through \(P_{n} \)). Lower levels capture fine-grained temporal details, while higher levels encode abstract, long-term dependencies. To enhance the interplay between visual and audio features, our network introduces two key components: (I) a \textbf{Audio \& Visual-guided Adapters} and (II) an \textbf{Adaptive Pooling Module}. The audio-guided adapter enriches visual features with acoustic context, while the visual-guided adapter enhances audio representations with visual semantics. These are followed by the adaptive pooling module that refines cross-modal integration by dynamically adjusting feature importance across different scales, ensuring optimal weighting of short- and long-duration event cues. 

\subsection{Modality-guided Adapters:}
These adapted modality-guided adapters are strategically positioned after the top-down or bottom-up fusion processes to leverage information from both modalities for mutual enrichment. For the audio-guided adapter features, let \( A_l \in \mathbb{R}^{T_{A} \times d} \) represent the audio embeddings and \( V_l \in \mathbb{R}^{T_{V} \times d} \) (where \( l \in \{1,2,3,4,5,6\} \)) denote the visual features in different scales. We then apply max-sigmoid attention to integrate the audio features into the visual features:

\begin{equation}
V_l' = V_l \cdot \delta \left( \text{max}_{j \in \{1..C_A\}} \left( V_l A_j^\top \right) \right)^\top
\end{equation}
where the  \( \delta \) represents the sigmoid activation function.
% \[
% \delta(x) = \frac{1}{1 + e^{-x}}
% \]

Similarly, for the visual-guided adapter, we apply max-sigmoid attention to integrate visual features into audio features:

\begin{equation}
A_l' = A_l \cdot \delta \left( \text{max}_{k \in \{1..C_V\}} \left( A_l V_k^\top \right) \right)^\top
\end{equation}

In both equations, \( V_l' \) and \( A_l' \) represent the updated feature maps for video and audio, respectively, which are then concatenated with their corresponding cross-stage features to produce the final outputs. Note that \( C_A \) and \( C_V \) represent the number of audio and video embedding channels, respectively.

\begin{figure}[t]
    \centering
    \includegraphics[trim={8.9cm 10cm 10cm 0.5cm}, height=2.9cm]{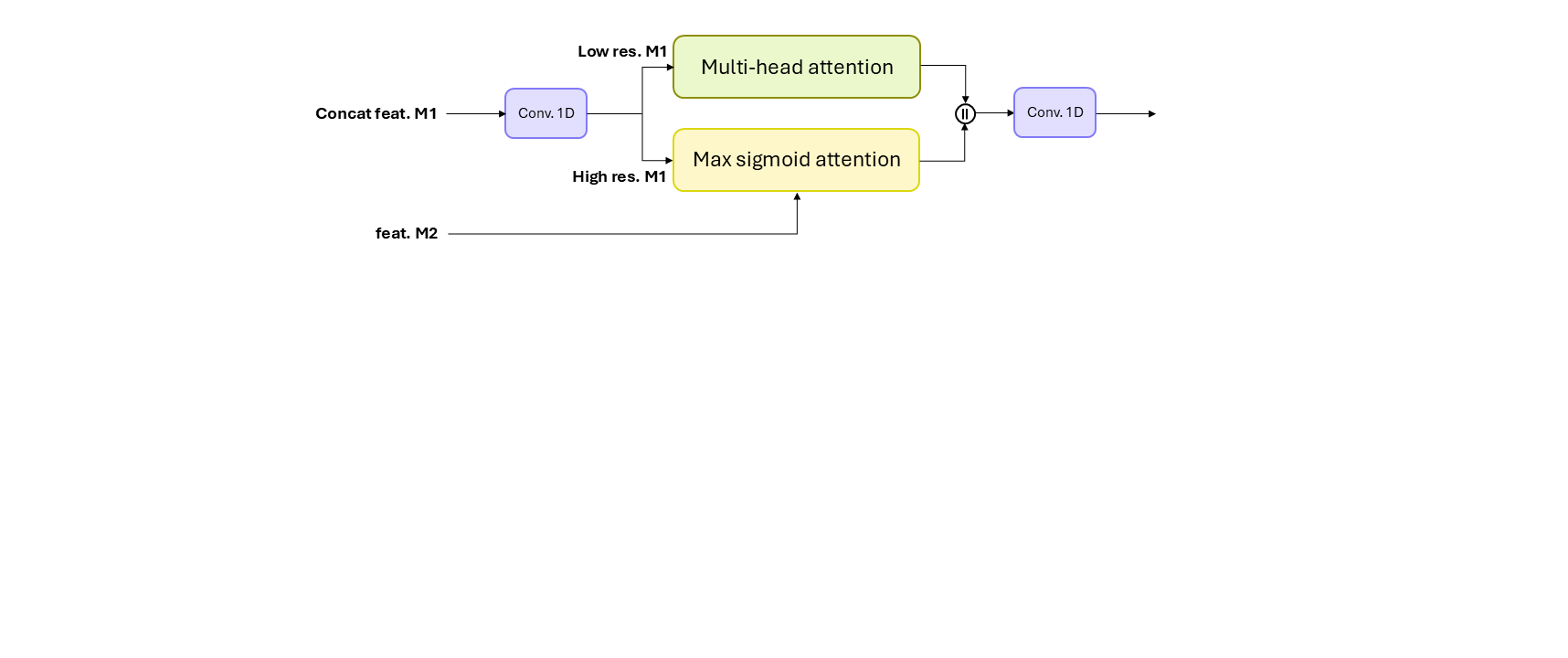}
      \vspace{-20pt}

    \caption{Detailed diagram of the max sigmoid module. This module is a key component of our path aggregation network, facilitating adaptive feature fusion between modalities. It takes high- and low-resolution feature maps from modality M1 and combines them with features from modality M2, using a max sigmoid function to dynamically select the most relevant features for enhanced representation learning.
    }
    \label{fig:maxsig}
\end{figure}

\subsection{Adaptive Pooling Module (APM)}
To further enhance the video features with audio information, the APM aggregates multi-scale audio features into \(N\) temporal segments, resulting in audio tokens \(\tilde{A} \in \mathbb{R}^{N \times d}\). 
We aggregate multi-scale audio features into $N$ temporal segments to establish a shared temporal resolution between audio and video modalities. Since the extracted audio and video features may have different temporal resolutions due to variations in sampling rates and processing pipelines, directly aligning them can be challenging. By fixing both modalities to $N$ segments, we enforce a structured representation where each token captures a comparable temporal extent across both modalities, ensuring better temporal synchronization between audio and video. This normalization reduces modality mismatches and allows multi-head attention to operate more effectively by aligning the most relevant audio cues with corresponding video segments. Additionally, this strategy helps manage computational efficiency by reducing the number of tokens processed while maintaining rich cross-modal interactions.
These tokens are then used to update the video embeddings \(V\) through multi-head attention:

\begin{equation}
V' = V + \text{MultiHead-Attention}(V, \tilde{A}, \tilde{A})
\end{equation}

Conversely, for audio enhancement, the APM aggregates multi-scale video features into \(N\) spatial regions, producing video tokens \(\tilde{V} \in \mathbb{R}^{N \times d}\). These tokens update the audio embeddings \(A\):

\begin{equation}
A' = A + \text{MultiHead-Attention}(A, \tilde{V}, \tilde{V})
\end{equation}

This approach allows for efficient integration of audio context into video representations and vice versa. The resulting updated embeddings, \(V'\) and \(A'\), offer richer cross-modal representations that improve the overall performance of our audio-visual fusion network. 

% \section{More Qualitative Results}
% Fig.~\ref{fig:sup} provides additional qualitative results, illustrating the predictions of our model variants across different input modalities.
% The figure illustrates the benefits of cross-modal perception for event localization. 
% \begin{figure}[tb]
%   \centering
% \includegraphics[trim={1.7cm 0cm 0cm 0cm},width=1.11\linewidth]{sec/Sup.png}
  
%   \caption{Additional qualitative results on the UnAV-100 test set. GT represents the ground truth, A denotes predictions from the audio-only model, V corresponds to the visual-only model’s predictions, and AV indicates the predictions made by our audio-visual model.}
%   \label{fig:sup}
% \end{figure}

\subsection{Limitations and Future Works}
While DEL demonstrates strong performance in dense audio-visual event localization, certain limitations remain. First, like many previous approaches~\cite{geng2023dense,zhang2022actionformer,lin2019bmn, fish2023multi}, the model depends on pre-extracted features. Exploring end-to-end trainable feature extractors could further enhance performance. Second, DEL primarily focuses on audio-visual fusion but does not explicitly incorporate language-based cues, which could provide richer contextual understanding, especially for complex multi-event scenarios. Finally, expanding the framework to handle unseen categories through few-shot or self-supervised learning techniques could improve its robustness and generalization in open-world settings.

\end{document}